# A Novel Approach For Finger Vein Verification Based on Self-Taught Learning

Mohsen Fayyaz[1], Masoud PourReza[2]
[1]Malek-Ashtar University of Technology
[2]Semnan University
Tehran, Iran
[1]mohsen.fayyaz89@gmail.com
[2]pourreza.masoud@gmail.com

Mohammad Hajizadeh Saffar[1], Mohammad Sabokrou[2], Mahmood Fathy[3]
[1,2]Malek-Ashtar University of Technology
[3]Iran University of Science and technology
Tehran, Iran
[1]hajizadeh.m@gmail.com

*Abstract*—In this paper, we propose a method for user Finger Vein Authentication (FVA) as a biometric system. Using the discriminative features for classifying theses finger veins is one of the main tips that make difference in related works, Thus we propose to learn a set of representative features, based on auto-encoders. We model the user finger vein using a Gaussian distribution. Experimental results show that our algorithm perform like a state-of-the-art on SDUMLA-HMT benchmark.

## I. INTRODUCTION

Biometric technology plays a key role in authentication systems. Biometric authentication systems verify the identity of people using their biometrics features. There are important biometric properties for biometrics features to be used in personal authentication process. Some of the most important properties are: uniqueness, universality, and permanence [1]. Finger veins are situated inside the body and because of this natural property, it is hard to forge and spoof them. Another key property of finger vein pattern authentication is the assurance of aliveness of the person, whose biometrics are being proved.

Finger vein pattern authentication process consist of four steps: (1) data acquisition, (2) preprocessing, (3) feature extraction, and (4) classification.

In data acquisition step, vein images are captured using infrared scanner. Captured images are affected by influence of blood pressure of veins, body temperature, and environment circumstances. To be more accurate, preprocessing step is necessary.

Preprocessing step consists of following stages: Extracting region of interest (ROI), background removing, and enhancement. For extracting ROI and removing background, some morphological operations can be done. In [2], ROI has been extracted by using interphalangeal joints. In enhancement stage, many methods have been exploited. In [3], improved adaptive Niblack threshold segmentation algorithm has been employed.

To classify users based on their finger vein, feature extraction methods are exploited (e.g. Skeleton, Hough transform)

As classical handcrafted low-level features, such as HOG and HOF, skeleton, hough transform may not be universally suitable and discriminative enough, We propose a FVA system based on unsupervised feature learning method. This method has used an autoencoder to learn best features for representing finger vein images. Using autoencoder not only enhanced the feature extraction process, but also made the system heavy preprocessing needless.

The rest of this paper is organized as follows: we survey the works related to FVA. Proposed method is introduced in section III. This section includes the feature representation and the classification method. Experimental results and their comparisons have been described in section IV. Section V concludes this paper and propose some suggestions for future work.

## II. RELATED WORK

The research area of biometrics is gaining more attention recently. Finger vein based authentication system is interested because of the finger vein intrinsic properties. As mentioned, FVA process consists of (1) preprocessing, (2) feature extraction, and (3) classification.

In preprocessing step, for enhancing veins in images various filters and transforms are used. In [4] Gabor filter has been used for vein enhancement. Curvelets transform and Steerable filter [2] also can be used for enhancement. Being negligent at local shape of a vein is one of the issues of vein tracking. Restoration algorithms which minimize the scattering effect of IR image acquisition have been used for enhancement In [5].

Many feature extraction methods are used in this research area such as, SIFT [6], minutiae [7], statistical measures [8], local binary pattern [9]. In [10] skeleton method have been introduced. Recognizing objects and patterns using skeleton features is not deformation invariant. There are two main problems in FVA systems: (1) low quality of infrared images and (2) pose variation of the finger. For overcoming to these problems, manifold learning has been used in [11].

For classifying task, each finger vein image must compare against reference finger vein models which are learned based a bit reference finger vein for each user. (There are one model equivalent to each user) The new sample will be checked with all models to identify it.

Selecting "good" threshold is important parameter in authentication systems for accepting or rejecting a finger vein image, which means that how much the input pattern should be similar to the reference pattern. Consequently, choosing the best threshold is one of the crucial steps. There are two types of thresholds: (1) global and (2) local. In global threshold, system will choose one threshold value for all users. Although in local threshold, system must choose one threshold per user. However respect to generality and complexity the global threshold has a better performance rather than local threshold, but in comparison of accuracy term it is worse [12].

### III. PROPOSED METHOD

Feature learning, is one of the new fields of machine learning that recently has considered by many researchers [13]. Feature learning is trying to define discriminative features and learn them unsupervised. As we have not enough labeled samples for each user, the Self-thought learning method is used. The main advantage of self-taught learning is the ability to use unlabeled data in a supervised classification [14].

Autoencoder is one of the unsupervised learning methods that has been proposed in the field of self-taught learning. An autoencoder tries to learn the best value for weights of hidden layer to set the output equal to the input layer. The autoencoder has been applied to diverse types of problems [15, 16]. Figure 1 shows the architecture of an autoencoder with one hidden layer.

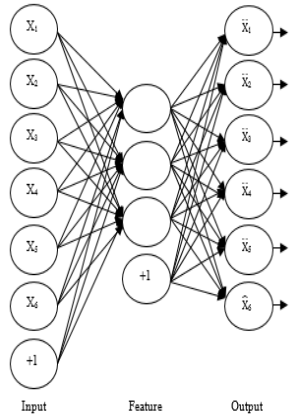

Figure 1 Architecture of Autoencoder

Suppose $x$ is the set of input features. To learn features from input data, the basic autoencoder with regularization term to prevent over-fitting, attempts reconstructing input features by minimizing following cost function. In order to learn features that are more effective and having a sparser set of represented features, the sparsity constraint can impose on the autoencoder network (Eq. 1-3):

$$J(W,b) = \arg\min_{W,b} \frac{1}{m} \sum_i \|h_{w,b}(x^{(i)}) - x^{(i)}\|^2 + \lambda \sum_l \sum_{i,j}(W_{i,j}^l)^2 + \beta \sum_i KL(\rho||\hat{\rho}_j) \quad (1)$$

$$KL(\rho||\hat{\rho}_j) = \rho \log \frac{\rho}{\hat{\rho}_j} + (1-\rho)\log \frac{1-\rho}{1-\hat{\rho}_j} \quad (2)$$

$$\hat{\rho}_j = \frac{1}{m}\sum_i [a_j^2(x^{(i)})] \quad (3)$$

Where $W$ is weight matrix mapping nodes of each layer to next layer nodes, and $b \in \mathbb{R}$ is a bias vector. $KL(\rho||\hat{\rho}_j)$ is the Kullback-Leibler (*KL*) divergence between a Bernoulli random variable with mean $\rho$ and a Bernoulli random variable with mean $\hat{\rho}_j$, which is the average activation of hidden unit $j$. The $x^{(i)}$ is the $i$-th training example, $\beta$ is the weight of the sparsity penalty term, m is the number of training set and finally $\lambda$ is weight decay parameter.

Choosing the best set of discriminative features for extracting from input data is one of the most challenging aspects of designing a verification system. For better classification, these features should be selected carefully to make data separable in their n-dimensional space. The best set of features for separating data can be achieved using an autoencoder and unsupervised learning method [12, 14]. The proposed method consist of three main phases: Feature Learning, Classification and Verification (Figure 2).

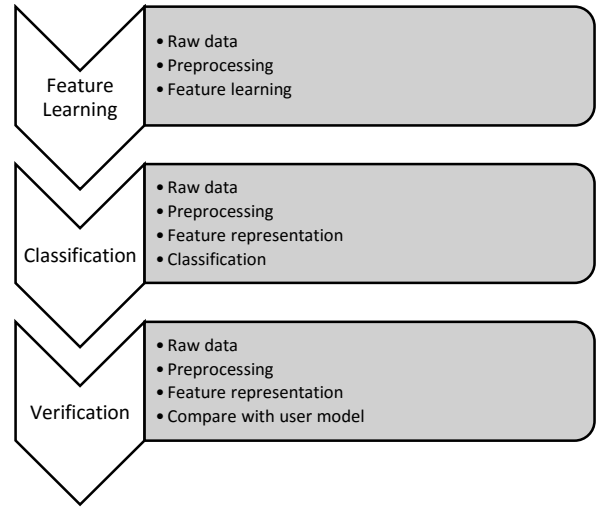

Figure 2 Proposed System Architecture

In the first phase, input raw data were given to an autoencoder to learn the features. Then the classification process for each user was performed with the set of learned features and as a result, a model for each user has been achieved. Finally, test data have been compared with the created model for each user to calculate the system accuracy.

There are three main steps in this architecture:

### A. Preprocessing

Preprocessing is the first step of system procedure after receiving the input data. The task of this step is to convert raw data to the aligned and acceptable input for autoencoder. For this reason, the dimensions of input images have been justified and the mean of images have become zero for normalization. After that, a contrast enhancement based on histogram remapping has been used. For finding the best histogram reference for remapping, we have used a genetic algorithm. This genetic algorithm is based on histogram remapping to produce natural looking images and the goal is clarifying the veins in output images. The chromosome structure has been defined based on [17]. The fitness function for evaluating each chromosome has been shown in equation 5.

$$fittness(x) = \log(\log(E(I(x))) * edges(I(x)) \quad (5)$$

Where $I(x)$ is the enhanced image and $edges(I(x))$ is the number of detected edges by sobel edge detector. In this equation, sum of intensity values has been shown by $E(I(x))$ and a log-log measure has been used to prevent producing unnatural images [17]. Figure 3 shows some output of genetic algorithm for histogram remapping.

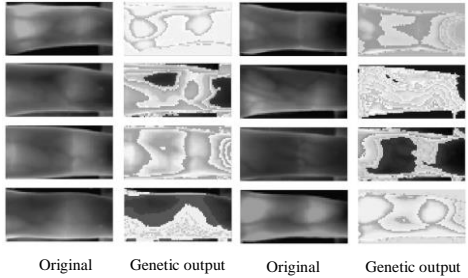

Original    Genetic output    Original    Genetic output

Figure 3 Sample of genetic output of histogram remapping

In addition, the correlation of adjacent pixels in images can cause some problems in the process of training and as a result, some of extracted features may became useless. Therefore, Principal Component Analysis (PCA) and Whitening algorithms were applied on images to reduce the size of input data and decrease the correlation of adjacent pixels. In addition, Whitening can make data less redundant and the features become the same variance.

### B. Feature Learning

As mentioned, a sparse autoencoder based on gradient descent has been used for feature extraction.

In this system, firstly input images have been divided to small patches and the autoencoder has learned the features from these patches. Secondly, the learned features have been convolved with the input image. Finally, the mean pooling process has been exploited on convolved features to achieve pooled features. These pooled convolved features have been used for classification phase.

### C. Classification

As the number of training samples are very limited, and the creating a model without enough training samples can meet "under fitting", So, two approach are applied in this paper: (1) Selecting a suitable classifier (2) representing the raw data using autoencoder, sparsely.

We use a one-class classifier as the user finger veins have been set for being target class and its outlier class has been set by other user's finger vein images. Based on experimental results, a Gaussian classifier with auto-optimized parameters has been used.

### IV. EXPRIMENTAL RESULTS

The proposed system has been evaluated using SDUMLA-HMT[1] Finger Vein database. It is the first free finger vein dataset. This dataset consists of 100 peoples' fingers veins images. In the capturing process, each subject has been asked to provide images of his/her index finger, middle finger and ring finger of both hands, and the collection for each of the 6 fingers is repeated for 6 times.

We categorized this dataset into three subsets for autoencoder training (feature learning), creating a model for each user by a one-class classifier and evaluation. For training the autoencoder, all images except the index finger of right hand of people, which have number of 3000 have been used. The images of people's right hand index finger have been used for model creation and evaluation which have number of 600.

Feature extraction has done by using a sparse autoencoder. This autoencoder comprises one hidden layer with 4000 nodes and the limited Broyden–Fletcher–Goldfarb-Shanno algorithm (L-BFGS) method with 700 iteration for minimization function. Figure 4 shows the illustration of the features that have been learned.

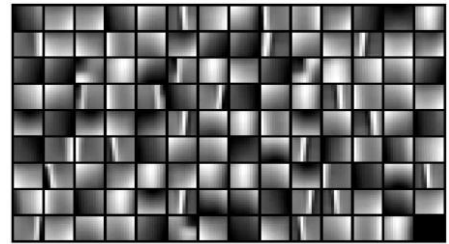

Figure 4 Illustration of learned features

The efficiency of the autoencoders have been tested based on the ability of the learned features to separate represented peoples' finger vein images from each other. To test such ability, a Gaussian classifier has been trained using represented features of tree images from each person's right hand index finger. The other tree images of right hand index fingers have been used to test the classifier. Test procedure has been done based on 10-fold class validation. To evaluate the classification process, the Equal Error Rate (EER) and Area Under the Cure (AUC) have been calculated. .

. For tuning the parameters of autoencoder such as number of iteration and the size of hidden layers, an experiment has been designed. Base on this experiment, the size of hidden layer varies from 1000 to 4000 and the iteration of autoencoder varies from 100 to 700. EER and AUC results have been shown in table 2 and table 3, respectively.

Table 1 EER experimental results for different hidden size and iterations

| Iteration/Hidden size | 1000 | 2000 | 3000 | 4000 |
|---|---|---|---|---|
| 100 | 1.26 | 1.66 | 1.33 | 1.63 |
| 200 | 1.48 | 1.49 | 1.31 | 1.33 |
| 300 | 1.70 | 1.33 | 1.30 | 1.03 |
| 400 | 1.60 | 1.34 | 1.20 | 1.14 |
| 500 | 1.50 | 1.36 | 1.10 | 1.26 |
| 600 | 1.71 | 1.54 | 1.06 | 0.98 |
| 700 | 1.93 | 1.73 | 1.03 | 0.70 |

---

[1] It is available at http://mla.sdu.edu.cn/sdumla-hmt.html

Table 2 AUC experimental results for different hidden size and iterations

| Iteration/Hidden size | 1000 | 2000 | 3000 | 4000 |
|---|---|---|---|---|
| 100 | 99.34 | 99.07 | 99.35 | 99.20 |
| 200 | 99.22 | 99.20 | 99.35 | 99.35 |
| 300 | 99.11 | 99.34 | 99.36 | 99.50 |
| 400 | 99.13 | 99.34 | 99.33 | 99.39 |
| 500 | 99.15 | 99.35 | 99.30 | 99.29 |
| 600 | 99.02 | 99.22 | 99.42 | 99.48 |
| 700 | 98.89 | 99.10 | 99.55 | 99.67 |

Due to change mitigation in more than 700 iterations, the iteration value has been set to 700. The rate of enhancement of EER and AUC decreased for hidden sizes larger than 4000 while computational costs increased and had been prone to curse of dimensionality. Finally, the size 4000 has been chosen because of its computational efficiency and appropriate accuracy.

As a comparison between the proposed method and other approaches, Table 4 shows the EER reported for some related works.

Table 3 Compare proposed method with related works

| Method | EER (%) |
|---|---|
| Yu, et al. [18] | 0.761 |
| Liu, et al. [19] | 0.8 |
| Khalil-Hani and Lee [20] | 0.87 |
| **Proposed Method** | **0.70** |

Table 4 indicates that proposed method have the best performance in comparison with competing algorithms. This method's EER is 0.70 percent, where the next best method is 0.761 percent reported for the method described by Yu, et al. [18].

## V. CONCLUTION

In this paper, a finger vein verification system has been proposed based on feature representation. In contrast with other works in this field that use hand-crafted features for their datasets, this system uses self-taught feature learning. Therefore, the best set of features have been extracted from the dataset and the result with the high accuracy has been achieved. In addition, according to the automation of the feature extraction and learning process, this method is dataset independence. The other contribution of the proposed system is having high accuracy in classification without doing any special preprocessing. In addition, because of using simple Gaussian classifier, the testing phase has a low computation cost and a high computing time efficiency.

As future work, more preprocesses can be implemented to improve the sharpness of finger vein images. In addition, the proposed system has been designed with one hidden layer and the effect of increasing the layers of autoencoder on system accuracy can be experimented.